\documentclass{article}

\usepackage{arxiv}

\usepackage[utf8]{inputenc} 
\usepackage[T1]{fontenc}    
\usepackage{hyperref}       
\usepackage{url}            
\usepackage{booktabs}       
\usepackage{amsfonts}       
\usepackage{nicefrac}       
\usepackage{microtype}      
\usepackage{lipsum}
\usepackage{multirow}       
\usepackage{graphicx}       
\usepackage{subcaption}     
\usepackage{caption}        

\title{Vilio: State-of-the-art Visio-Linguistic Models applied to Hateful Memes}

\author{
  Niklas Muennighoff \\
  Peking University \\
  \texttt{https://github.com/Muennighoff} \\
  \texttt{muennighoff@stu.pku.edu.cn} \\
}

\begin{document}
\maketitle

\begin{abstract}
This work presents Vilio, an implementation of state-of-the-art visio-linguistic models and their application to the Hateful Memes Dataset. The implemented models have been fitted into a uniform code-base and altered to yield better performance. The goal of Vilio is to provide a user-friendly starting point for any visio-linguistic problem. An ensemble of 5 different V+L models implemented in Vilio achieves 2nd place in the Hateful Memes Challenge out of 3,300 participants. The code is available at \url{https://github.com/Muennighoff/vilio}.

\end{abstract}


\section{Introduction}
Language always appears as part of a different medium, such as written text in vision or spoken text in sound. For some problems, extracting the language only suffices for useful machine learning models. Pure language models have therefore had considerable impact on our  lives in areas such as machine translation, text classification tasks or language reasoning problems \cite{brown2005language}. Often, however, the underlying medium is relevant and must be understood in conjunction with the language. \\
One such area is Internet memes. They combine an image and a superimposed text to provide a nuanced message. Often the image or text on its own is not enough to convey the message. That nuanced message can be hateful, but manual classification and removal of hateful memes is costly for social platforms. The Hateful Memes Challenge \cite{kiela2020hateful} proposed by Facebook AI and hosted on the DrivenData platform aims to leverage machine learning models to solve this problem. \\
Current state-of-the-art Vision+Language machine learning models are based on the transformer architecture \cite{vaswani2017attention}. Among them, there are two prevalent approaches:
\textbf{Single-stream models}, such as VisualBERT \cite{li2019visualbert}, UNITER \cite{chen2019uniter}, OSCAR \cite{li2020oscar}, use a single transformer to process the image and language input at the same time.
\textbf{Dual-stream models}, such as LXMERT \cite{tan2019lxmert}, ERNIE-ViL \cite{yu2020ernie}, DeVLBERT \cite{zhang2020devlbert}, VilBERT \cite{lu2019vilbert}, rely on separate transformers for vision and language, which are then combined towards the end of the model. \\
This work contributes the following:
\begin{itemize}
    \item A code base of 12 different vision+language models organized like huggingfaces transformers library \cite{wolf2019huggingface} to promote future V+L research
    \item Performance-enhancing modifications applied to state-of-the-art V+L models
    \item An evaluation of different models on the Hateful Memes Dataset \cite{kiela2020hateful} \& necessary code for future research on the dataset
\end{itemize}

\section{Problem Statement}
\label{sec:headings}

\begin{figure}[t]
\begin{subfigure}{0.5\textwidth}
\centering
\includegraphics[height=4cm]{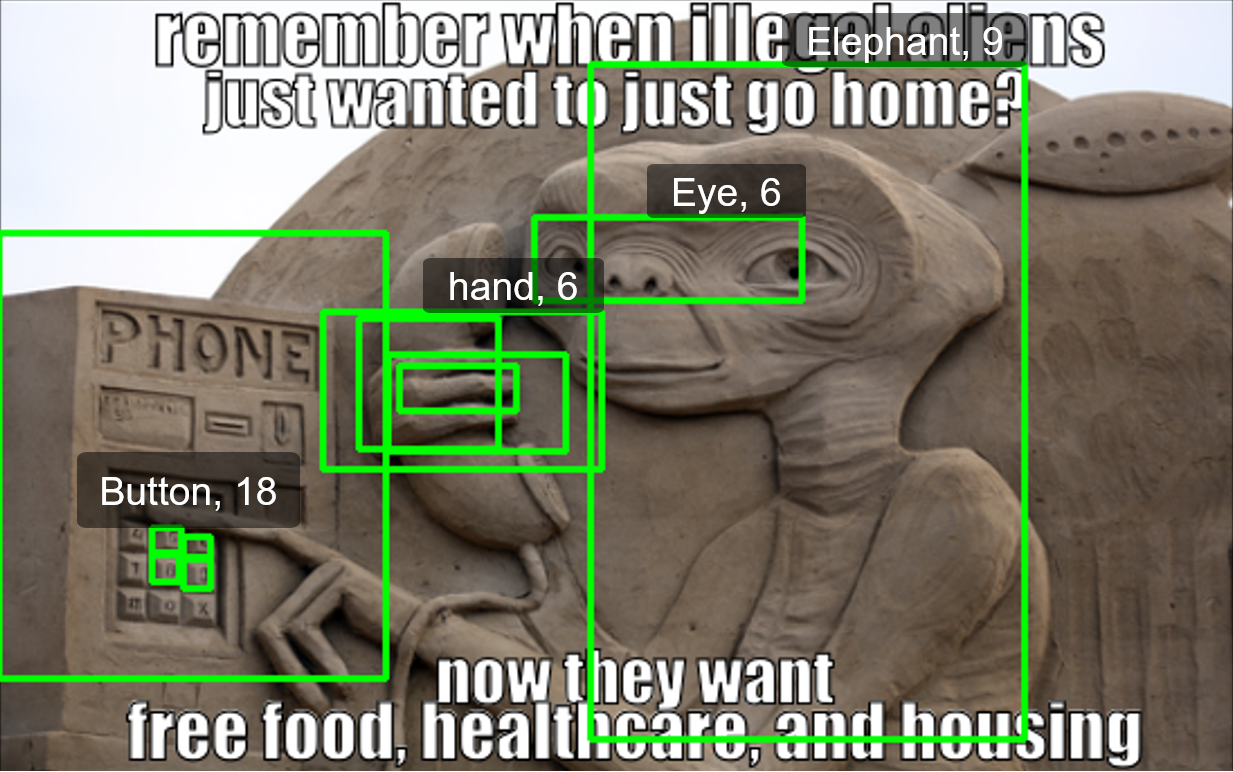} 
\caption{Hateful meme}
\label{fig:subim1}
\end{subfigure}
\begin{subfigure}{0.5\textwidth}
\centering
\includegraphics[height=4cm]{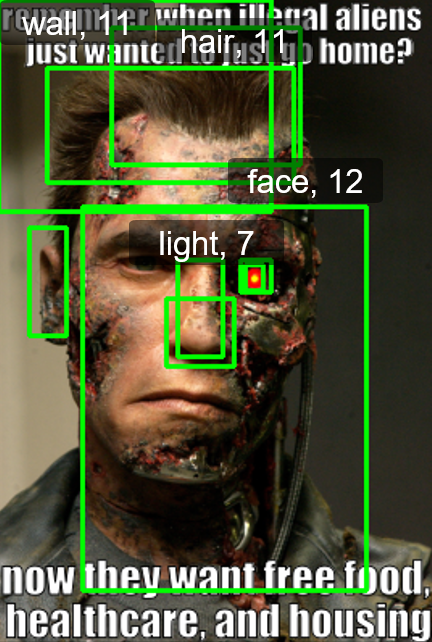} 
\caption{Non-hateful meme}
\label{fig:subim2}
\end{subfigure}
\captionsetup{justification=centering}
\caption{Predicted RoIs \& the top four predicted objects on a confounder example from the Hateful Memes Dataset. The hateful meme has been neutralized by removing \& replacing E.T. with an unrelated image. \textit{©Getty Images}}
\label{fig:fig1}
\end{figure}

The Hateful Memes Dataset \cite{kiela2020hateful} consists of a training set of 8500 images, a dev set of 500 images \& a test set of 1000 images. The meme text is present on the images, but also provided in additional jsonl files. 
To increase its difficulty, the dataset includes text- \& vision confounders. Such confounders change from being hateful to non-hateful or vice-versa by swapping either text or image only. Figure \ref{fig:fig1} is one such example. They ensure that models must reason about both vision \& language. A vision-only or language-only model cannot succeed in the task.\\
Participants in the competition submitted a probability for the likelihood of a meme being hateful for each of the 1000 test images. The submitted probabilities were then used to calculate the area under the curve of the receiver operating characteristic:
\begin{equation}
AUROC =  \int_{x=0}^{1} \mbox{TPR}(\mbox{FPR}^{-1}(x)) \, dx
\end{equation}
The AUROC score was used as the competitions key metric. Intuitively, it penalizes models that are bad at ordering memes by hatefulness. Whether the probability values itself are high or low does not matter, only how they are ranked.

\section{Approach}
\label{sec:others}

\begin{figure}[t]
    \centering
    \includegraphics[width=14cm]{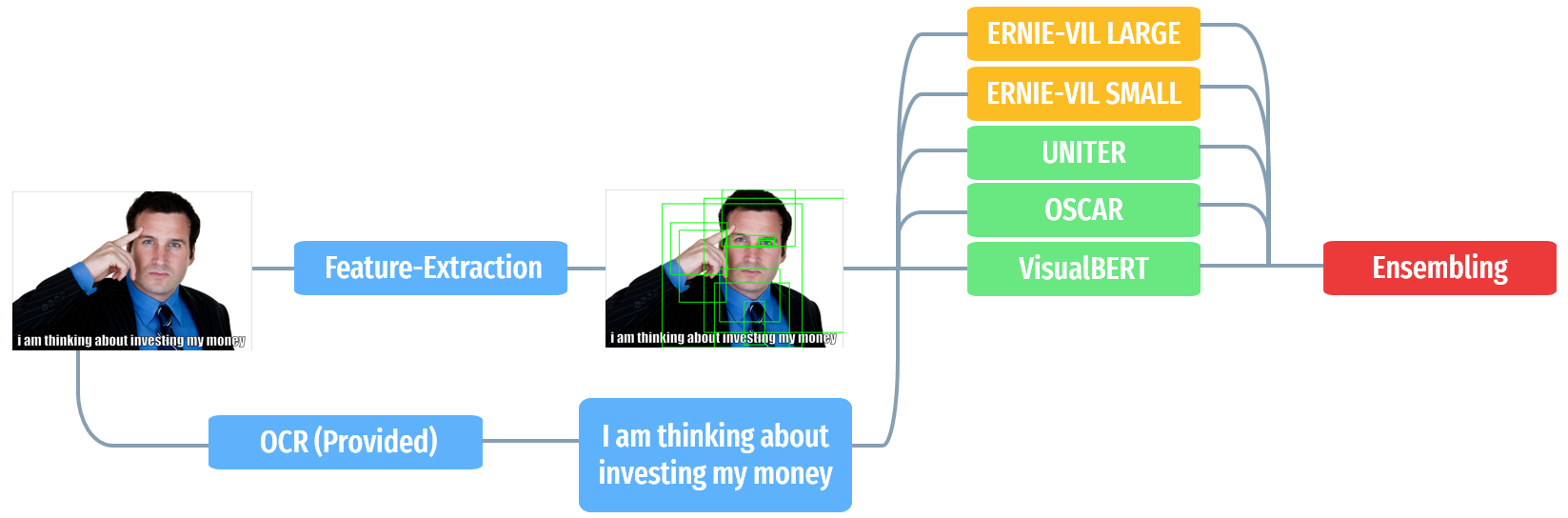}
    \captionsetup{justification=centering}
    \caption{Pipeline on the Hateful Memes Dataset using Vilio split into three stages: Preparation, Modelling \& Ensembling \textit{©Getty Images}}
    \label{fig:fig2}
\end{figure}

A high-level overview is present in Figure \ref{fig:fig2}. The pipeline consisted of three stages: Preparation, Modelling \& Ensembling.

\subsection{Preparation}

In a first step, features were extracted from images using the detectron2 framework \cite{wu2019detectron2}. Using features instead of whole images speeds up the training process and still captures the most relevant content. It has been shown that using diverse features can boost performance \cite{jiang2018pythia}. Therefore, different models are used for feature extraction, pre-trained on VisualGenome with \& without attributes \cite{Anderson2017up-down}. In addition, varying Regions of Interest (RoIs) are kept in each feature extraction. Min- \& Maxboxes are set to the same number either 36, 50, 72 or 100. Figure \ref{fig:fig1} shows an example of predicted RoI's and the four most commonly predicted objects on two memes. \\
Together with the meme text, which has been extracted using optical character recognition (OCR) and provided in the dataset, features are then fed into the models.

\subsection{Modelling}

The following provides an overview of general settings and changes made to individual models that were used in the final solution to the Hateful Memes Challenge. The code repository also provides additional models that were not used in the submission. 

\subsubsection{General Settings}

Pre-trained weights provided by the original authors of the V+L models are used. VisualBERT \& OSCAR are also task-specific pre-trained on the Hateful Memes Dataset. All models are then finetuned using binary cross-entropy loss and a batch size of 8. The Adam optimizer \cite{kingma2019method} is used with a learning rate of 1e-5 and 10\% linear warm-up steps \cite{ma2019adequacy}. Gradients are clipped at 5 for VisualBERT, OSCAR \& UNITER and at 1 for ERNIE-ViL. VisualBERT, OSCAR \& UNITER are trained for 5 epochs and Stochastic Weight Averaging \cite{izmailov2018averaging} is used during the last 25\% of training. ERNIE-ViL models are trained for 5000 steps. The weights from the last step are taken for all models and used for inference on the test set. \\
Overall, not much time was spent on hyperparameter optimization, as fundamental architecture changes have a larger impact.

\subsubsection{ERNIE-ViL}

The ERNIE-ViL model by PaddlePaddle \cite{yu2020ernie} is based on VilBERT \cite{lu2019vilbert} and the ERNIE transformer \cite{zhang2019ernie}. In addition to extracted features, the original ERNIE-ViL is pre-trained on ground-truth labelled boxes. As there are no ground-truth boxes provided for the Hateful Memes Dataset, a second set of features extracted with 10-100 boxes is used as a fake ground-truth. In addition to pre-trained weights on Conceptual Captions (CC) \cite{sharma2018conceptual}, task-specific pre-trained weights on VCR \cite{zellers2019recognition} are used to increase diversity. 

\subsubsection{UNITER \& OSCAR}

Updates are made to both UNITER \& OSCAR to reflect changes in the transformers library \cite{wolf2019huggingface}, such as an updated activation function and embedding calculations. OSCAR is also task-specific pre-trained on Hateful Memes using Image-Text-Matching (ITM) and Masked Language Modelling (MLM). Adding the predicted objects \& attributes during task-specific pre-training as in the original implementation was not beneficial. The classifier from LXMERT with GeLU activation \cite{tan2019lxmert} is used.
The OSCAR \& UNITER pre-trained weights are based on the BERT transformer \cite{devlin2018bert}. The code to use RoBERTa \cite{liu2019roberta} and other language transformers is provided in Vilio. However, without the pre-training, they cannot match performance.

\subsubsection{VisualBERT}

Similar to UNITER \& OSCAR, VisualBERT was updated based on the transformers library as of September, 2020. Like OSCAR, VisualBERT model is task-specific pre-trained using MLM. While the original VisualBERT uses the same token type ids for the language \& visual input, the Vilio implementation creates a separate visual token type. Therefore, token type weights are reinitialized and retrained from scratch. This improved the model by an absolute 1.2\% on the AUROC metric. Multi-sample dropout \cite{inoue2019multi} and learnt weights for averaging of the transformer layers further improve the model. A linear classification head is used with 500 times the learning rate of the transformer backbone.

\subsection{Ensembling}

For each of the models 3-5 seeds with different extracted features are averaged. Subsequently, the averaged predictions of each model are fed into an ensembling loop that applies Simple Averaging, Rank Averaging, Power Averaging \& Simplex Optimization \cite{wu2005triopt} to produce a final submission. The weights for the Simplex Optimization are learned based on dev set predictions and then applied to test set predictions trained on the full dataset (train+dev). 

\section{Results \& Discussion}

\begin{table}
  \centering
  \begin{tabular}{ p{4cm} p{4cm} p{2cm} p{2cm} } 
    \toprule
          &     & \textbf{Validation} & \textbf{Test} \\ [0.5ex] 
    Source & Model & AUROC & AUROC \\ [0.5ex]
    \midrule
    \multirow{5}{8em}{Hateful Memes Baseline} & Human & - & 82.65 \\ 
    \cmidrule(r){2-4}
    & ViLBERT & 71.13 & 70.45 \\
    & VisualBERT & 70.60 & 71.33 \\
    & ViLBERT CC & 70.07 & 70.03 \\
    & VisualBERT COCO & 73.97 & 71.41 \\ [0.5ex] 
    \midrule
    \multirow{6}{8em}{Vilio} & VisualBERT & 75.49 & 75.75 \\ 
    & OSCAR & 77.16 & 77.30 \\
    & UNITER & 77.75 & 78.65 \\
    & ERNIE-ViL Base & 78.18 & 77.02 \\
    & ERNIE-ViL Large & 78.76 & 80.59 \\
    \cmidrule(r){2-4}
    & Ensemble & \textbf{81.56} & \textbf{82.52} \\ [0.5ex]
    \bottomrule
  \end{tabular}
  \newline
  \newline
  \caption{Model performance.}
  \label{tab:table}
\end{table}


\subsection{Performance \& Limits}
\label{sec:performance}

The individual models performance on the validation set and test set can be seen in Table \ref{tab:table}. When ensembled, Vilio effectively closes the gap between baseline models \& human performance on Hateful Memes. The organizers have noted, however, that the humans tested on were not experts and that the actual human performance might be closer to 100\%. Investigating this and establishing a new human baseline makes for an interesting future direction. \\
While not reported, the ensemble accuracy of Vilio using absolute labels, not probabilities, is only 75.40\% compared to 84.70\% for humans on the test set \cite{kiela2020hateful}. The reason for this is that Vilio has been optimized for ranking, but humans are better at binary predictions than probabilities. Accuracy leaves for a considerably large gap to close. \\
Including the averaged seeds, the Vilio ensemble is made up of 19 trained models. Especially the ERNIE-ViL models tend to be instable, hence five seeds are averaged. Adding Stochastic Weight Averaging \cite{izmailov2018averaging} for ERNIE-ViL, not just the other models, could help tackle this issue. Achieving the same results with less seeds and less compute is something worth looking into.

\subsection{Future Work}

With the framework laid down, directions for future work are vast. Apart from the suggestion in Section \ref{sec:performance}, four ideas are:

\begin{itemize}
    \item Can we solve hateful GIFs? 
    \item In the Hateful Memes dataset meme captions were standardized, but in reality they may vary in font and size. This is useful information that may help the model determine hatefulness. Can we therefore integrate the OCR algorithm into the trainable model? 
    \item With recent advances in applying transformers to vision \cite{dosovitskiy2020image}, can we skip the feature extraction and apply single-stream or dual-stream transformers from scratch?
    \item Single-stream encoders, such as VisualBERT, seem to outperform dual-stream encoders, such as VilBERT, with the exception of ERNIE-ViL. ERNIE-ViL, which copies VilBERT's encoders, differentiates itself with changes like its Scene Graph Parser {\cite{yu2020ernie}}. Could an ERNIE-VisualBERT model with a single-stream encoder outperform the dual-stream ERNIE-ViL? 
\end{itemize}


\section{Acknowledgements}

I would like to thank Greg Lipstein, Casey Fitzpatrick \& the others from DrivenData for hosting the competition and their patience. I also want to thank Facebook AI, especially Douwe Kiela, Hamed Firooz \& Tony Nelli, for making this competition possible in the first place.
Lastly, I'm very grateful for the insightful discussions I had with Liunian Harold Li, Hao Tan \& Antonio Mendoza. Without your help, it would have been a lot more difficult. Thanks a lot!


\bibliographystyle{unsrt}  
\bibliography{references} 

\end{document}